\pdfoutput=1

\documentclass[11pt]{article}

\usepackage{authblk}
\usepackage[preprint]{acl}

\usepackage{times}
\usepackage{latexsym}

\usepackage[T1]{fontenc}

\usepackage{microtype}

\usepackage{inconsolata}

\usepackage[T1]{fontenc}    
\usepackage{hyperref}       
\usepackage{url}            
\usepackage{booktabs}       
\usepackage{amsfonts}       
\usepackage{microtype}      
\usepackage{color}
\usepackage{algorithm,algpseudocode}
\usepackage{caption}
\usepackage{mwe}
\usepackage{lipsum}

\usepackage{algorithm}
\usepackage{algpseudocode}

\usepackage{graphicx}
\usepackage{amsmath}
\usepackage{amssymb}
\usepackage{float}
\usepackage{stfloats}
\usepackage{bbm}
\usepackage{dsfont}
\usepackage{algorithm}
\usepackage{array}
\usepackage[caption=false,font=normalsize,labelfont=sf,textfont=sf]{subfig}
\usepackage{textcomp}

\usepackage{verbatim}
\usepackage{multirow}
\usepackage{graphicx}
\usepackage{pifont}
\usepackage{color}
\usepackage[hang,flushmargin]{footmisc} 
\usepackage{booktabs}
\usepackage[normalem]{ulem}

\usepackage{soul}
\usepackage{float}
\usepackage{tcolorbox}
\usepackage{tikz}

\soulregister{\cite}7
\soulregister{\ref}7
\soulregister\eqref7

\usepackage{colortbl}

\definecolor{azure(web)(azuremist)}{rgb}{0.94, 1.0, 1.0}
\definecolor{oldlace}{rgb}{0.99, 0.96, 0.9}
\definecolor{pearl}{rgb}{0.94, 0.92, 0.84}
\definecolor{seashell}{rgb}{1.0, 0.96, 0.93}
\definecolor{silver}{rgb}{0.75, 0.75, 0.75}
\definecolor{platinum}{rgb}{0.9, 0.89, 0.89}
\definecolor{almond}{rgb}{0.94, 0.87, 0.8}
\definecolor{lightskyblue}{RGB}{173, 216, 230}

\usepackage{arydshln}

\title{\textsc{DAST}: Difficulty-Aware Self-Training on Large Language Models}

\author[1]{\bf Boyang Xue}
\author[2,$^{*}$]{\bf Qi Zhu}
\author[1]{\bf Hongru Wang}
\author[1]{\bf Rui Wang}
\author[3]{\bf Sheng Wang}
\author[4]{\bf Hongling Xu}
\author[2]{\\ \bf Fei Mi}
\author[2]{\bf Yasheng Wang}
\author[2]{\bf Lifeng Shang}
\author[2]{\bf Qun Liu}
\author[1,\thanks{~~Co-corresponding authors.}]{\bf Kam-Fai Wong}

\affil[1]{The Chinese University of Hong Kong, $^{\mathrm 2}$Huawei Noah’s Ark Lab}
\affil[3]{The University of Hong Kong}
\affil[4]{Harbin Institute of Technology, Shenzhen}

\begin{document}
\maketitle

\begin{abstract}

Present Large Language Models (LLM) self-training methods always under-sample on challenging queries, leading to inadequate learning on difficult problems which limits LLMs' ability.
Therefore, this work proposes a difficulty-aware self-training (DAST) framework that focuses on improving both the quantity and quality of self-generated responses on challenging queries during self-training.
DAST is specified in three components: 1) sampling-based difficulty level estimation, 2) difficulty-aware data augmentation, and 3) the self-training algorithm using SFT and DPO respectively.
Experiments on mathematical tasks demonstrate the effectiveness and generalization of DAST, highlighting the critical role of difficulty-aware strategies in advancing LLM self-training.

\end{abstract}

\section{Introduction}
\label{sec:intro}

\begin{quote}
    \textit{What doesn’t kill you makes you stronger.}
    \hspace*{\fill}\textit{— Friedrich Wilhelm Nietzsche}
\end{quote}

The lack of extensive, high-quality human-curated training data for Large Language Models (LLMs) constrains the potential upper bounds of their capacities, particularly on complex reasoning tasks \citep{cobbe2021training}.
Recently, self-training techniques of LLMs have garnered increasing attention, which iteratively fine-tunes LLMs on their self-generated outputs, attaining sustained improvements and diminishing the reliance on human interventions
\citep{gulcehre2023reinforcedselftrainingrestlanguage,singh2024humandatascalingselftraining,huang-etal-2023-large,zelikman2022starbootstrappingreasoningreasoning}.

\begin{figure*}[ht!]
    \centering
    \includegraphics[width=0.97\textwidth]{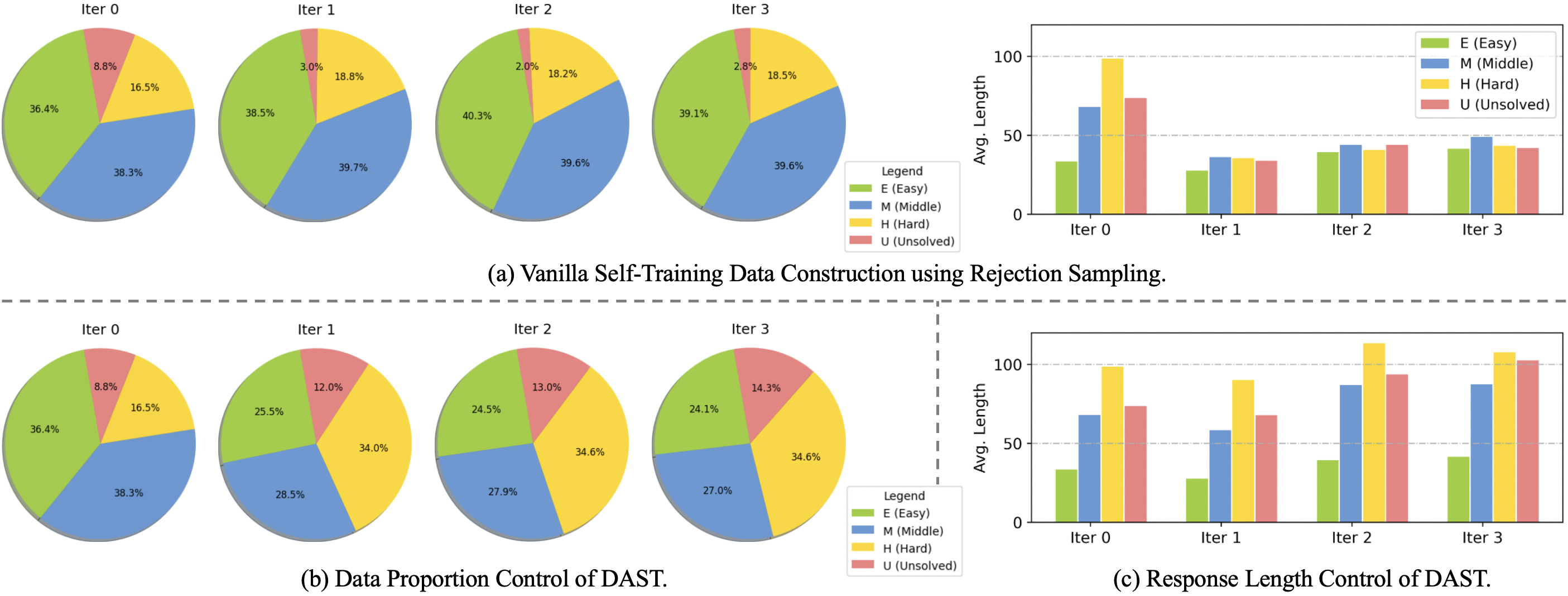}
    \caption{Changes of data proportion and response length distribution of samples in different difficulty levels during a three-round self-training process.
    The vanilla rejection sampling to construct training data (a) is widely employed in \citet{singh2024humandatascalingselftraining,gulcehre2023reinforcedselftrainingrestlanguage,sordoni2023joint,zelikman2022starbootstrappingreasoningreasoning}.
    (b) and (c) are the proposed DAST aim to control data proportion and response lengths for challenging queries.
    Note that in iteration 0, the training data $\mathcal D_{\mathrm u}$ is the original dataset $\mathcal D_{\mathrm o}$ with ground-truth labels, while during iteration 1, 2, and 3, the training data is combined of self-generated data $\mathcal D_{\mathrm a}$ and the original dataset $\mathcal D_{\mathrm o}$.
    All the difficulty levels are measured on the initial policy $\mathcal M_0$ on the GSM8K test set and are fixed during self-training.}
    \label{fig:difficulty}
\end{figure*}

To ensure the quality of LLMs' self-generated training data, previous works employ rejection sampling \citep{sordoni2023joint} to filter out low-quality or incorrect responses with external reward models \citep{gulcehre2023reinforcedselftrainingrestlanguage} or ground-truth labels \citep{singh2024humandatascalingselftraining}. 
This may lead to LLM over-sampling originally adept simple queries while under-sampling challenging queries \citep{ding2024mitigating,tong_dart-math_2024}. 
LLMs' insufficient learning in challenging instances is primarily in two aspects during self-training.
First, when fixing the sampling number, only a few even or no correct responses are acquired on challenging queries, which iteratively exacerbates the distribution imbalance of the training data and severely overfitting on simple questions (Left hand of Figure \ref{fig:difficulty} (a)).
Second, the lengths of sampled self-generated responses on difficult questions are not enough (Right hand of Figure \ref{fig:difficulty} (a)).
Given that challenging problems require more thinking steps \citep{snell2024scalingllmtesttimecompute,damani2024learninghardthinkinputadaptive}, the quality of these responses tends to be lower.
As a result, LLMs can not adequately learn from challenging tasks, thereby restricting their capacity improvements.

Considering the above two issues, this work proposes a \textbf{d}ifficulty-\textbf{a}ware \textbf{s}elf-\textbf{t}raining (DAST) framework which focuses on increasing both the quantity and quality of self-generated responses on challenging queries during self-training:
1) DAST employs a sampling-based, model-specific method to estimate the difficulty level of each query.
2) Two data augmentation approaches are employed to balance the distribution and improve the response quality of training data given the difficulty levels.
Specifically, we perform up-sampling on challenging questions to control the data proportion of different difficulty levels.
We also employ a difficulty-matched few-shot prompting method to control the lengths of responses, encouraging LLMs to increase thinking steps on challenging questions.
These two methods are combined incrementally. 
3) We finally iteratively perform the above difficulty estimation and data augmentation steps in several rounds for LLM self-training using supervised fine-tuning (SFT) and direct preference optimization (DPO) \citep{NEURIPS2023_a85b405e} respectively.

Experiments are conducted on both the in-domain and out-of-domain tasks on various mathematical datasets.
Results demonstrate that \textsc{DAST} significantly enhances LLMs' math ability and generalizability over several baselines.

Our contributions are as follows: 1) This work first comprehensively incorporates difficulty level into LLM self-training, demonstrating the significance of considering difficulty for future works. {Our codes are available on \href{https://github.com/AmourWaltz/DAST}{\texttt{https://github.com/AmourWaltz/DAST}}}; 
2) We propose two data augmentation methods in DAST to improve both quantity and quality on challenging queries using the estimated difficulty level;
3) We conduct experiments and validate that DAST can enhance LLM's math ability and generalizability using SFT and DPO respectively.

\section{DAST Framework}
\label{sec:method}

\subsection{Difficulty Level Estimation}

We employ a sampling-based, model-specific method to estimate the difficulty level of each question to the model. 
Given the initial policy $\mathcal M_0$ and the training set $\mathcal D_{\mathrm{o}}={\left \{ \boldsymbol x_i, \boldsymbol {\hat r}_i, \boldsymbol {\hat y}_i \right \}}_{i=1}^{N}$, where $\boldsymbol x_i, \boldsymbol r_i, \boldsymbol y_i$ represent the question, rationale, and the ground-truth answer respectively.
Each rationale $\boldsymbol {\hat r}_i = \left [\hat r_{i,1}, \dots,\hat r_{i,l} \right ]$ contains $l$ reasoning steps where $l$ varies in $\boldsymbol {\hat r}_i$.
For each $(\boldsymbol x_i, \boldsymbol {\hat r}_i, \boldsymbol {\hat y}_i)$ and a prompt set $\mathcal P$ containing $K$ different few-shot prompts, 
we employ each few-shot exemplar $\boldsymbol{p}_{k}\in \mathcal P$ with the question $\boldsymbol x_i$ for the policy $\mathcal M_{0}$ to generate the $k$-th response $\left ({\boldsymbol y_i}^{(k)},{\boldsymbol r_i}^{(k)}\right )=\mathcal M_{0}\left (\boldsymbol{p}_{k}, \boldsymbol x_i\right )$ using temperature sampling ($T=0.2, \mathrm{top}~p=0.9$).
We obtain the response set $\boldsymbol Y_{i}=\{{\boldsymbol y_i}^{(k)}\}_{k=1}^{K}$ and the label set $\boldsymbol Z_i = {\left \{ {z_{i}}^{(k)} \right \}}_{k=1}^{K}$ by comparing each extracted answers in $\boldsymbol Y_{i}$ with the ground-truth $\boldsymbol{\hat y}_i$ to determine the correctness
(${z_{i}}^{(k)}\in\left \{ 0,1 \right \}$, 1 for \textit{True} and 0 for \textit{False}).
The difficulty level $d_i$ is estimated as follows \footnote{In this study, the challenging queries refer to the queries estimated in difficulty levels of Middle, Hard, and Unsolved}.
Details and splits of four difficulty levels are in Table \ref{table:diff}.

\begin{align}
\label{eq:diff}
    d_i=P\left (\boldsymbol Y_i|\boldsymbol x_i\right)=\frac{\sum_{k=1}^{K}\mathbb I\left ({\boldsymbol y_i}^{(k)}={\boldsymbol{\hat y}}_i\right)}{K}
\end{align}

\subsection{Data Augmentation}

We augment $\mathcal D_{\mathrm o}$ with the strategy $\mathcal A(\cdot)$ for each query $\boldsymbol x_i$ according to $d_i$ by controlling the data proportion and response lengths on $\mathcal M$ to obtain an augmented dataset $\mathcal D_{\mathrm a}$ for self-training as follows.

\paragraph{Data Proportion Control}
As in the left hand of Figure \ref{fig:difficulty} (a), the construction of self-training data using rejection sampling may bias simple questions.
Therefore, we set different sampling numbers $K$ for different difficulty levels $d_i$ of $\boldsymbol x_i$.
More specifically, the sampling number $K$ will multiply by a coefficient $\beta$ determined by $d_i$ as presented in Table \ref{table:diff}.
For $d_i\in \{M, H, U\}$ which indicates that $\boldsymbol x_i$ is a challenging question, $\beta$ is larger to increase the number of correct responses sampled from the policy $\mathcal M$.
The sampled responses will be added into $\mathcal{D}_{\mathrm a}$.
As illustrated in Figure \ref{fig:difficulty} (b), we can dynamically control the proportion of samples in all difficulty levels and balance the distribution of the training data in each self-training iteration.

\paragraph{Response Length Control}
As in the right hand of Figure \ref{fig:difficulty} (a), the lengths of responses generated using the vanilla few-shot sampling method are in averaged length for all difficulty levels during self-training (iterations 1, 2, and 3) and relatively shorter than lengths of the ground-truth responses in $\mathcal D_{\mathrm o}$ (iteration 0).
To generate lengthy and difficulty-matched responses, we propose a difficulty-matched few-shot (DMFS) prompting method: for each difficulty level $d\in\{E, M, H, U\}$, we select samples from the training set that exceed the average response length of this difficulty level to construct four prompt sets $\mathcal P_{E}, \mathcal P_{M}. \mathcal P_{H}, \mathcal P_{U}$.
DMFS examples are employed based on $d_i$ to sample responses for $\boldsymbol{x}_i$ on $\mathcal M$.
Sampled responses will be added into $\mathcal D_{\mathrm a}$.
Therefore, length distribution of $\mathcal D_{\mathrm a}$ is close to the ground truth in iteration 0 as in Figure \ref{fig:difficulty} (c), which improves the response quality with more thinking steps \citep{snell2024scalingllmtesttimecompute,yeo2025demystifyinglongchainofthoughtreasoning}.



\begin{algorithm}[t!]
    \fontsize{10pt}{11pt}\selectfont
    \caption{\textsc{DAST} Algorithm}
    \begin{algorithmic}[1]
         \State \textbf{Input:} Training set $\mathcal D_{\mathrm{o}}$, validation set $\mathcal D_{\mathrm{v}}$, number of iterations $\mathcal T$, policy model at $t$-th iteration $\mathcal M_{t}$.
        \State \textbf{Output:} Optimized policy $\pi_{\theta}$.
        \For{$t = 1$ to $\mathcal T$}
            \For{$i = 1$ to $\left | \mathcal D_{\mathrm{o}}\right |$}
                \State Estimate difficulty level $d_i$ of $\boldsymbol x_i$
                \State Obtain ${\{ \boldsymbol r_i^{(m)},\boldsymbol y_i^{(m)} \}}_{m=1}^{M}=\mathcal A(\boldsymbol x_i, d_i)$
                \For{$\boldsymbol y_i = \boldsymbol y_i^{(1)}$ to $\boldsymbol y_i^{(M)}$}
                    \If{$\boldsymbol y_i \equiv \boldsymbol {\hat y}_i$}
                        \State Label and add $(\boldsymbol x_i, \boldsymbol r_i^{+}, \boldsymbol y_i^{+})$ to $\mathcal D_a^{(t)}$
                    \Else
                        \State Label and add $(\boldsymbol x_i, \boldsymbol r^{-}_i, \boldsymbol y^{-}_i)$ to $\mathcal D_{a}^{(t)}$
                        \EndIf 
                    \EndFor
                \EndFor
            \State Update training set $\mathcal D_{\mathrm{u}}=\mathcal D_{\mathrm{o}}\cup \mathcal D_a^{(t)}$
            \While{$\mathcal M_{t-1}$'s accuracy improves on $\mathcal D_v$}
                \State Optimize $\mathcal M_{t-1}$ on $\mathcal D_{\mathrm{u}}$ using SFT or DPO by minimizing $\mathcal L_{\mathrm{sft}}/\mathcal L_{\mathrm{dpo}}$ as in Equation \ref{eq:sft} or \ref{eq:dpo}
                \EndWhile
            \State $\mathcal M_{t}\leftarrow \mathcal M_{t-1}$
        \EndFor
    \end{algorithmic}
    \label{dast:algorithm}
\end{algorithm}

\subsection{Self-Training}

As presented in Algorithm \ref{dast:algorithm}, in the $t$-th iteration, the training set $\mathcal D_{\mathrm{u}}$ is updated by merging the augmented dataset $\mathcal D_{\mathrm{a}}^{(t)}$ and initial training set $\mathcal D_{\mathrm{t}}$, ensuring $\mathcal D_{\mathrm{u}}$ doesn't diverge too much from $\mathcal D_{\mathrm{t}}$.
The policy $\mathcal M_j$ is fine-tuned based on $\mathcal M_{j-1}/\mathcal M_{0}$ on $\mathcal D_u$ using SFT$/$DPO \citep{NEURIPS2023_a85b405e} by optimizing $\mathcal L_{\mathrm{sft}}/\mathcal L_{\mathrm{dpo}}$ in Equation \ref{eq:sft}$/$\ref{eq:dpo} respectively.
$\mathcal M_j$ is trained to be converged while the accuracy doesn't increase on the validation set $\mathcal D_{\mathrm v}$.
Specifically, we denote DAST using SFT/DPO by \textbf{DAST-S}/\textbf{DAST-D}.
For DAST-S, we investigate only employing data proportion control or length control, and denote by \textbf{DAST-P} and \textbf{DAST-L} respectively.


\section{Experimental Setting}
\label{sec:exp}




\paragraph{Datasets}
During the training stage, we jointly combine training sets from \textbf{GSM8K} \citep{cobbe2021training} and \textbf{MATH} \citep{hendrycks2021math} as $\mathcal D_{\mathrm t}$.
We evaluate \textbf{in-domain (ID)} performance on the corresponding test sets.
We also assess the \textbf{out-of-domain (OOD)} performance three challenging test sets: \textbf{TAL-SCQ} \citep{matheval2023tal} \textbf{College} \citep{tang2024mathscale}, and \textbf{TheoremQA} \citep{chen2016training}.
We standardize the data format as in Appendix \ref{appendix:prompt} and employ the evaluation script of MWPBench \footnote{\href{https://github.com/microsoft/unilm/tree/master/mathscale}{https://github.com/microsoft/unilm/tree/master/mathscale}} \citep{tang2024mathscale} to judge the correctness of the extracted answer compared with the ground-truth label.
Dataset details are in Appendix \ref{appendix:dataset}.


\paragraph{Baselines}
We utilize in-context learning (\textbf{ICL}) \citep{brown2020language} to generate responses.
We also employ several SFT-based and DPO-based baselines.
SFT-based baselines include: 1) single-round standard \textbf{SFT} and difficulty-aware rejection tuning (\textbf{DART}) \citep{tong_dart-math_2024} (specified in \textbf{DART}-\textit{Uniform} and \textbf{DART}-\textit{Prob2Diff}); and 2) multi-round \textbf{ReST-EM} \citep{singh2024humandatascalingselftraining}.
DPO-based \citep{NEURIPS2023_a85b405e} baselines include single- and multi-round DPO (\textbf{DPO} and \textbf{mDPO}).
Detailed implementations of the above baselines can be referred to Appendix \ref{appendix:baseline}.



\section{Results and Analysis}
\label{sec:result}

\begin{figure*}[!ht]
  \centering
  \includegraphics[width=0.99\textwidth]{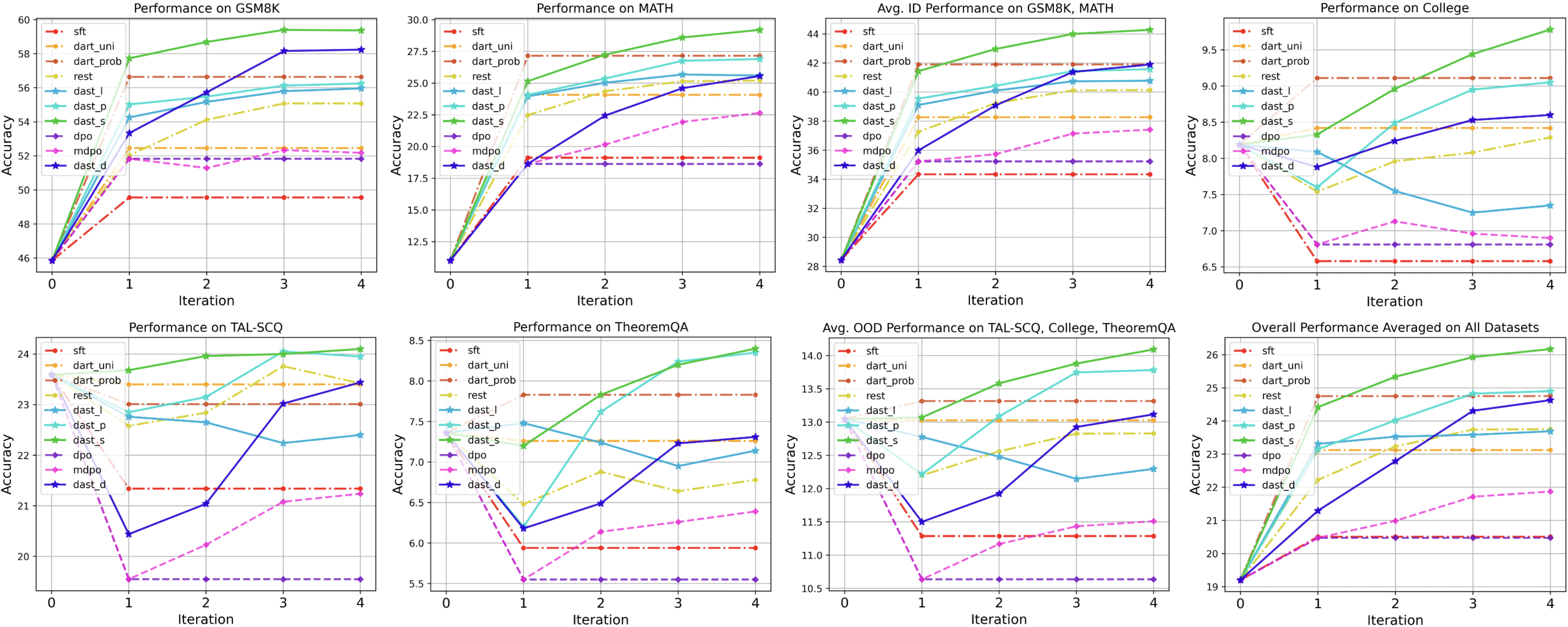}
    \caption{Performance results of DAST over various baselines on both in-domain (ID) and out-of-domain (OOD) mathematical test sets using Llama-3.1.
    Note that the names of employed baselines are in lowercase.}
  \label{fig:main_exp}
\end{figure*}

\subsection{Main Experiments}

Experiments are conducted on {\textbf{{Llama-3.1-8B}}} ({Llama-3.1}) \citep{llama3modelcard} in this work.
As in Figure \ref{fig:main_exp}, several findings can be found below.

1. With different sizes of self-training data in each iteration, \textbf{DAST-S and DAST-D consistently yield superior performance over corresponding SFT and DPO baselines with comparable or less data}, exhibiting the effectiveness and efficiency of DAST for both SFT and DPO during self-training.
Data size statistics are presented in Table \ref{table:main_exp}.

2. DAST-P exhibits better performance compared to DAST-L, suggesting that \textbf{increasing the data size can gain more improvements than increasing the response lengths for challenging queries}. 
This can be attributed to that the initial policy is suboptimal and the sampled lengthy responses are also low-quality.
Therefore, raising the data quantity can lead to more obvious gains.

3. \textbf{DAST-S and DAST-P can better generalize to OOD tasks than others}.
DAST enables LLMs to adequately learn more diverse challenging questions, thereby achieving more pronounced improvements in relatively challenging OOD tasks.

\subsection{Effects of Data Proportion Control}

\begin{figure}[!ht]
  \centering
  \includegraphics[width=0.49\textwidth]{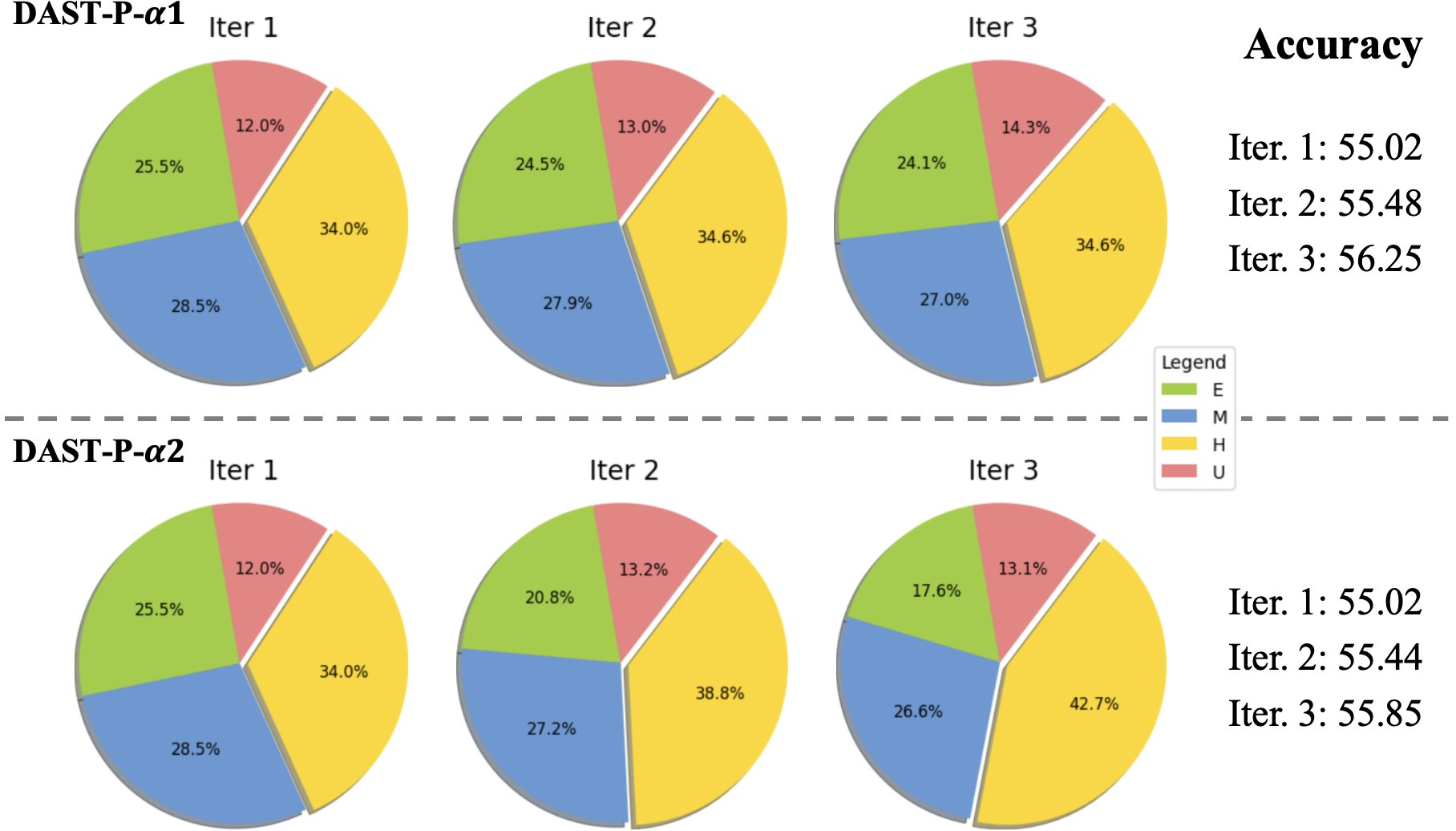}
    \caption{Results of data proportion control.}
  \label{fig:dast_p}
\end{figure}

In this part, we investigate the research question \textit{"As self-training progresses iteratively, will increasing the proportion of challenging samples lead to further improvements?"}.
We control the proportions of challenging queries with fixed data size in each iteration by adjusting $\beta$ during self-training as illustrated in Figure \ref{fig:dast_p}.
Results suggest that LLMs perform better when trained on the dataset with a balanced distribution (DAST-P-$\alpha 1$) of different difficulty levels than more hard samples (DAST-P-$\alpha 2$) during self-training.
Excessive challenging samples may lead to a large distribution shift, affecting LLMs' original abilities on simple queries.

\subsection{Effects of Response Length Control}

In this part, we investigate the research question \textit{"Will the performance be further improved by employing difficult examples across all queries to generate lengthy responses during self-training?"}. We generate training data using few-shot examples from solely a single difficulty level in the first round of DAST to compare with our proposed difficulty-matched few-shot (DMFS) prompting method for sampling.
Results in Table \ref{table:length} suggest that training data generated by DMFS outperforms those obtained from any single level.
Tailoring response length to difficulty levels of queries is more effective, as sampling lengthy responses to simple queries may result in overthinking and undermine performances \citep{halawi2024overthinkingtruthunderstandinglanguage}.

\begin{table}[ht!]
    \centering
    \footnotesize
    \resizebox{.47\textwidth}{!}
    {\begin{tabular}{c|ccccc}
    \toprule
        \textbf{Exam. Level} & $E$ & $M$ & $H$ & $U$ & DMFS \\
        \hline
        \bf ID & 35.58 & 37.44 & 38.90 & 38.66 & \bf 41.94  \\
        \bf OOD & 11.45 & 12.15 & 12.48 & 12.06 & \bf 13.07 \\
    \bottomrule
    \end{tabular}}
    \caption{Results of response length control.}
\label{table:length}
\end{table}

\section{Conclusion}
\label{sec:conclu}

This work proposes a DAST framework to enhance both the quantity and quality of challenging queries during the self-training process, including three key parts: difficulty level estimation, data augmentation, and a self-training algorithm.
Experiments conducted on math tasks using SFT and DPO showcase the effectiveness and generalization of DAST.


\section*{Limitations}
\label{sec:limit}

The limitations of this work are as follows:

\paragraph{Response Quality}
This work enhances the response quality by solely increasing the thinking steps or lengths of responses.
Although improving response quality by adding length is simple yet effective for challenging queries, more explorations should be conducted to comprehensively evaluate the response quality in other dimensions.

\paragraph{Task Expansion}
Another limitation is that the experiments are solely conducted on mathematical reasoning tasks. 
This constraint primarily arises from that many tasks like long-form generations are also challenging to evaluate the generation quality.
Future research endeavors should prioritize a wider range of datasets of long-form generation tasks to thoroughly assess the applicability and effectiveness of DAST.

\section*{Acknowledgments}

\bibliography{custom}


\appendix

 

\section{Protocols}
\label{appendix:protocol}

\subsection{Definition of Notations}
\label{appendix:notation}

\begin{table*}[!ht]
  \centering
  \fontsize{10pt}{12pt}\selectfont
  {\begin{tabular}{cc}
    \hline
    \textbf{Notation} & \textbf{Description} \\
    \hline
    $\mathcal D_{\mathrm o}$ & Training set containing $N$ Question-Answering pairs. ($|\mathcal D_{\mathrm o}| = N$) \\
    $\mathcal D_{\mathrm v}$ & Validation set. \\
    $\mathcal P$ & Set of few-shot exemplars. \\
    $\mathcal M_{t}$ & Policy model in the $t$-th ieration where $\mathcal M_0$ is the initial policy. \\
    $\boldsymbol x_i$ & The $i$-th question sample. \\
    $\boldsymbol {\hat r}_i$ & The $i$-th ground-truth rationale path for $\boldsymbol x_i$. \\
    ${\boldsymbol {r}_i}^{(k)}$ & The $k$-th sampled rationale path to the $i$-th question $\boldsymbol x_i$. \\
    $\boldsymbol {\hat y}_i$ & The $i$-th ground-truth answer for $\boldsymbol x_i$. \\
    ${\boldsymbol {y}_i}^{(k)}$ & The $k$-th sampled response to the $i$-th question $\boldsymbol x_i$. \\
    $\boldsymbol {p}_k$ & $k$-th few-shot exemplar to sample ${\boldsymbol {y}_i}^{(k)}$. \\
    $K$ & Number of sampled responses. \\
    $\boldsymbol {Y}_i$ & Answering set containing $K$ sampled response $\left \{ {\boldsymbol {y}_i}^{(k)} \right \}$ for the $i$-th question $\boldsymbol x_i$. \\
    ${z_i}^{(k)}$ & The label of ${\boldsymbol {y}_i}^{(k)}$ (${z_{i}}^{(k)}\in\left \{ 0,1 \right \}$, 1 for \textit{True} and 0 for \textit{False}). \\
    $\boldsymbol {Z}_i$ & Label set corresponding to $\boldsymbol {Y}_i$. \\
    $\mathcal L_{\alpha}$ & Training loss functions SFT or DPO where $\alpha\in \left \{ \mathrm{sft}, \mathrm{dpo} \right \}$. \\
    $d_j$ & Estimated difficulty level for $\boldsymbol{x}$. \\
    $c$ & Co-efficient to control the data proportion of samples in different difficulty levels. \\
    $T$ & Temperature of sampling. \\
    $\mathcal T$ & Number of iterations. \\
    \hline
  \end{tabular}}
  \caption{Summarized notations in this work.}
\label{table:notation}
\end{table*}

The definitions of the notations in this work are summarized in Table \ref{table:notation}.

\subsection{Difficulty Level Split}
\label{appendix:difficulty}

\begin{table}[ht]
    \centering
    \small
    {\begin{tabular}{cccc}
    \toprule
        $p$ & Difficulty Level & Denotation $d_j$ & $\beta$ \\
        \hline
        [0.8, 1.0] & Easy & $E$ & 1 \\
        $[0.4, 0.8)$ & Middle & $M$ & 3 \\
        (0.0, 0.4) & Hard & $H$ & 5 \\
        0.0 & Unsolved & $U$ & 5 \\
    \bottomrule
    \end{tabular}}
    \caption{Difficulty level split.}
\label{table:diff}
\end{table}

\subsection{Equations}

SFT is optimized by minimizing the negative log-likelihood loss as follows.

\begin{align}
\label{eq:sft}
    \mathcal L_{\mathrm{sft}}=\mathbb E \left [-\log \mathcal M_{j-1}(\boldsymbol y_i^{+}, \boldsymbol r_i^{+}|\boldsymbol x) \right ]
\end{align}

DPO is optimized to minimize the preference loss as follows.

\begin{align}
\label{eq:dpo}
    \mathcal L_{\mathrm{dpo}}={\mathbb E}\left [-\log \sigma \left (  \theta(\boldsymbol y_i^{+}, \boldsymbol r_i^{+}|\boldsymbol x) - \theta(\boldsymbol y_i^{-}, \boldsymbol r_i^{-}|\boldsymbol x) \right ) \right ]
\end{align}
where ${(\boldsymbol x_i, \boldsymbol y_i^{+}, \boldsymbol r_i^{+}, \boldsymbol y_i^{-}, \boldsymbol r_i^{-})\sim \mathcal D_{\mathrm u}}$ and $\theta(\cdot|\boldsymbol x)=\log \frac{\mathcal M_{j-1}(\cdot|\boldsymbol x)}{\mathcal M_{0}(\cdot|\boldsymbol x)}$.


\section{Related Works}
\label{sec:related}

\paragraph{LLM Self-Training}

LLM Self-Training \citep{gulcehre2023reinforcedselftrainingrestlanguage,singh2024humandatascalingselftraining} involves a machine learning paradigm where a LLM iteratively improves its performance by generating and leveraging its own synthetic data for further training without human intervention
also referring to self-taught \citep{zelikman2022starbootstrappingreasoningreasoning,hosseini2024vstartrainingverifiersselftaught}, self-evolving \citep{tao2024surveyselfevolutionlargelanguage}, or self-improve \citep{huang-etal-2023-large}.
Such self-training paradigms always involve a generation step by prompting LLMs to self-generate training data and an improve step by training the LLM on the self-generated data \citep{gulcehre2023reinforcedselftrainingrestlanguage}.
In the Generation step, to ensure the data quality, the generated data are always filtered and selected using rejection sampling \citep{yuan2023scalingrelationshiplearningmathematical} before being employed for training.
These signals can be reward scores returned by a reward model \citep{gulcehre2023reinforcedselftrainingrestlanguage}, the binary score to judge the correctness given gold answer for mathematical or coding tasks \citep{singh2024humandatascalingselftraining,yuan2023scalingrelationshiplearningmathematical,zelikman2022starbootstrappingreasoningreasoning,wang-etal-2024-self-training}, or two scores using two reward model for process and object respectively on reasoning tasks \citep{yang2024qwen25math}.
LLM itself can be also regarded as judge or the reward model \citep{yuan2024selfrewardinglanguagemodels, gu2025surveyllmasajudge}.

In the Improve step, the selected data are utilized to train the LLM using supervised fine-tuning (SFT) \citep{gulcehre2023reinforcedselftrainingrestlanguage,zelikman2022starbootstrappingreasoningreasoning,singh2024humandatascalingselftraining,xue2024ualignleveraginguncertaintyestimations} or reinforcement learning \citep{gulcehre2023reinforcedselftrainingrestlanguage,hosseini2024vstartrainingverifiersselftaught,wang-etal-2024-self-training}.
Some studies iteratively train the policy LLM based on the previously obtained LLM \citep{gulcehre2023reinforcedselftrainingrestlanguage} while some train the base LLM instead of the LLM obtained from the previous iteration \citep{wang-etal-2024-self-training,singh2024humandatascalingselftraining,zelikman2022starbootstrappingreasoningreasoning,wang-etal-2023-large,wang-etal-2024-enhancing}.

\paragraph{Data Synthesis on Math Problems}
Since the growth rate of high-quality data is significantly outpaced by the expansion of training datasets, synthetic data has emerged as a promising solution \citep{wang2024surveydatasynthesisaugmentation} to address the data capacity limitation and further improve LLM performance according to scaling laws \citep{kaplan2020scalinglawsneurallanguage}.
Self-training paradigm employs LLM itself to generate the synthetic training data on mathematical problems \citep{singh2024humandatascalingselftraining,zelikman2022starbootstrappingreasoningreasoning,wang-etal-2024-self-training}.
\citet{tong_dart-math_2024} proposes to synthesize more responses for challenging questions.
\citet{yu2024metamath} bootstraps the diversity of math problems by re-writing the training set and further fine-tunes LLM on the enhanced training set.
\citet{li-etal-2024-mugglemath} designs several re-writing principles to enhance both questions and responses to obtain an enhanced training set.
\citet{luo2025wizardmathempoweringmathematicalreasoning} proposes to synthesize more complex and diverse mathematical instructions to improve LLMs' mathematical reasoning ability.
\citet{ding2024mitigating} employs the Socratic-Guided Sampling (GSI) method to synthesize data to address the long-tail distribution issue during self-training.
Some studies also investigate synthesizing new questions \citep{huang2024keypointdrivendatasynthesisenhancement,zhou2024jiuzhang3}

Furthermore, Test Time Scaling Law \citep{snell2024scalingllmtesttimecompute} has attracted much attention recently, which proposed to consider allocating more computation resources in inference to generate high-quality responses.
These LLMs' self-generated data can be further used for LLM training to self-improve LLMs \citep{gulcehre2023reinforcedselftrainingrestlanguage}.
Many works validate that incorporating data multiply sampled on LLMs in inference can benefit LLMs and lead to further improvements such as dialogue system \citep{wang-etal-2023-retrieval,campagna2020zeroshottransferlearningsynthesized,wang2023surveyevolutionlanguagemodelbased}, multilingual LLMs \citep{saeki2024extendingmultilingualspeechsynthesis,xue2024comprehensivestudymultilingualconfidence}, and knowledge-intensive QA \citep{wang-etal-2024-role, puri2020trainingquestionansweringmodels, wang2025selfdcreasonactself}, which is a new trend for LLM training.
Although few additional computation costs are required, such works can still efficiently be utilized practically with significant improvements.

\section{Dataset Details}
\label{appendix:dataset}

\paragraph{GSM8K}
GSM8K \citep{cobbe2021training} \footnote{\href{https://github.com/openai/grade-school-math}{https://github.com/openai/grade-school-math}} is a high-quality multi-step mathematical reasoning dataset of diverse grade school math word problems constructed by human problem writers, including 7,472 training samples and 1,319 test samples.
All the questions take 2 to 8 steps to solve, involving a series of basic arithmetic operations to parse the final answer.

\paragraph{MATH}
MATH \citep{hendrycks2021math} \footnote{\href{https://github.com/hendrycks/math/}{https://github.com/hendrycks/math/}} is a challenging mathematical dataset with competition mathematics problems, consisting of 7,500 training samples and 5,000 test samples.
Each problem in MATH also has a full step-by-step solution which can be used to teach models to generate answer derivations and explanations across several subjects including algebra, geometry, number theory, counting and probability, calculus, etc.

\paragraph{TAL-SCQ}
TAL-SCQ5K-EN \citep{matheval2023tal} \footnote{\href{https://github.com/math-eval/TAL-SCQ5K}{https://github.com/math-eval/TAL-SCQ5K}} are high-quality mathematical competition datasets in English created by TAL Education Group with totally 5,000 samples. 
The TAL-SCQ dataset split 3,000 and 2,000 questions for training and testing respectively.
The questions are in the form of multiple-choice and cover mathematical topics at different levels of primary, junior high, and high school.
We format all the samples in standard QA format.

\paragraph{College} \citep{tang2024mathscale} \footnote{\href{https://github.com/microsoft/unilm/tree/master/mathscale/MWPBench}{https://github.com/microsoft/unilm/tree/master/mathscale/MWPBench}}
The College dataset contains 1281 training and 2818 test college-level mathematical problems extracted from 9 textbooks across 7 domains such as linear algebra and differential equations.
This dataset is to test generalization on complex mathematical reasoning in diverse domains.

\paragraph{TheoremQA} \citep{chen2016training} \footnote{\href{https://github.com/wenhuchen/TheoremQA}{https://github.com/wenhuchen/TheoremQA}}
The TheoremQA dataset contains 800 problems focused on utilizing mathematical theorems to solve challenging problems in fields such as math, physics, finance, and engineering, testing generalization on theoretical reasoning in general STEM.
The dataset is collected by human experts with very high quality.
We filter out the questions requiring pictures and remain 747 samples to test.

\section{Baseline Details}
\label{appendix:baseline}

\paragraph{ReST-EM}
Sampling Stage: Set the sampling temperature to 0.5. For each query, sample 10 responses.
Retain responses based on whether the final answer matches the ground truth.
Training Stage: Combine the sampled data from the current policy model with the original dataset $\mathcal D_{\mathrm o}$ to form a new training dataset, which is then used for supervised fine-tuning (SFT).

\paragraph{DAST-\textit{Uniform}}
Sampling Stage: Set the sampling temperature to 0.5.
During dataset construction, perform oversampling for difficult samples to ensure every sample has 4 correct responses.
Training Stage: Combine the sampled data with the original dataset $\mathcal D_{\mathrm o}$ to form a new training dataset, which is then used for supervised fine-tuning (SFT).

\paragraph{DAST-\textit{Prob2Diff}}
Sampling Stage: Set the sampling temperature to 0.5. During dataset construction, perform oversampling for difficult samples, applying a coefficient based on the difficulty level. More challenging samples are assigned more responses.
Training Stage: Combine the sampled data with the original dataset to form a new training dataset, which is then used for supervised fine-tuning (SFT).

\paragraph{DPO}
Sampling Stage: Set the sampling temperature to 0.5. The dataset construction is similar to SFT while we will also add negative samples into training data to conduct the DPO algorithm.

\paragraph{mDPO}
The sampling stage is similar to ReST-EM and we will also add negative samples into training data to conduct the DPO algorithm.
For the multi-round DPO, we sample the self-generated training data on the model obtained from the previous training iteration but we train the model from the initial policy as in Equation \ref{eq:dpo}.


\section{Prompt Template}
\label{appendix:prompt}

\begin{quote}
    \begin{tcolorbox}[colback=almond!20!white, colframe=almond!60!black, title=\textbf{Prompt and Problem Format}]
    \label{temp:icl}
        \small
        You are an excellent mathematician. Answer the following mathematical questions based on your knowledge. \\

        \#\#\# Question \#\#\#: \{\texttt{Question}\} \\
        \#\#\# Response \#\#\#: \\
        <think>\{\texttt{Reasoning steps}\}</think>. \\
        The answer is \textbackslash box\{\texttt{Answer}\}.

    \end{tcolorbox}
\end{quote}

\section{Implementation Details}
\label{appendix:training}

Experiments are conducted on {\textbf{{Llama-3.1-8B}}} ({Llama-3.1}) \footnote{\href{https://huggingface.co/meta-llama/Llama-3.1-8B}{https://huggingface.co/meta-llama/Llama-3.1-8B}} \citep{llama3modelcard}.

During dataset construction, we sample the responses using 8-shot examples by setting the sampling temperature to $T=0.5$.
For response length control of DAST, challenging samples are paired with longer few-shot examples.
When sampling, we will dynamically adjust the sampling number $K$ to control the training data in each iteration comparable as in Table \ref{table:main_exp}.

During training, ADAM parameter update is used in a mini-batch mode.
The initial learning rate of 1e-4 is utilized with the 0.05 warm-up ratio and 0.01 weight decay of the ADAM optimizer.
When training the models, we fix the training steps and ensure that all the models can be trained to convergences. 
Although the training data size of different methods are different, fixed training steps in total can maintain fairness for all the methods.

When decoding, the temperature is also set to 0.2 to be consistent with the sampling setting.
All the models are quantified using float16 (fp16) to load and save parameters.
The vLLM library \citep{kwon2023efficient} \footnote{\href{https://github.com/vllm-project/vllm}{https://github.com/vllm-project/vllm}} is utilized to accelerate the generation.
All the experiments are conducted on 4 $\times$ NVIDIA A100-40GB GPUs.

\begin{table}[t]
    \centering
    \small
    {\begin{tabular}{lcccccccccc}
    \toprule
        \multirow{1}{*}{\bf Method} & \multirow{1}{*}{\bf Iteration} & \multirow{1}{*}{\bf Data Size} \\
        \hline
        \hline
        \bf ICL & - & - \\
        \bf SFT & - & 15k \\
        {\bf DART}-\textit{Uniform} & - & 60k \\
        {\bf DART}-\textit{Prob2Diff} & - & 60k \\
        \hline
        \bf ReST-EM & 1 & 50k \\
        \bf ReST-EM & 2 & 55k \\
        \bf ReST-EM & 3 & 58k \\
        \bf ReST-EM & 4 & 58k \\
        \hline
        \bf DAST-P & 1 & 55k \\
        \bf DAST-P & 2 & 56k \\
        \bf DAST-P & 3 & 58k \\
        \bf DAST-P & 4 & 58k \\
        \hline
        \bf DAST-L & 1 & 56k \\
        \bf DAST-L & 2 & 56k \\
        \bf DAST-L & 3 & 56k \\
        \bf DAST-L & 4 & 56k \\
        \hline
        \bf DAST-S & 1 & 58k \\
        \bf DAST-S & 2 & 59k \\
        \bf DAST-S & 3 & 60k \\
        \bf DAST-S & 4 & 60k \\
        \hline
        \bf DPO & - & 15k \\
        \hline
        \bf mDPO & 1 & 50k \\
        \bf mDPO & 2 & 55k \\
        \bf mDPO & 3 & 58k \\
        \bf mDPO & 4 & 58k \\
        \hline
        \bf DPO-D & 1 & 58k \\
        \bf DPO-D & 2 & 59k \\
        \bf DPO-D & 3 & 60k \\
        \bf DPO-D & 4 & 60k \\
        \hline
    \bottomrule
    \end{tabular}}
    \caption{.}
\label{table:main_exp}
\end{table}

\end{document}